\documentclass[a4paper]{article}

\usepackage{INTERSPEECH2020}

\usepackage{spconf}
\usepackage{amsmath,setspace}
\usepackage{graphicx}
\usepackage{subcaption}
\usepackage{amssymb}
\usepackage{color}
\usepackage{bm}
\usepackage{bbm}
\usepackage[table]{xcolor}
\usepackage{url}
\usepackage{varwidth}
\usepackage{textcomp}   
\usepackage{gensymb}

\DeclareMathOperator{\E}{\mathbb{E}}

\title{Sentence level estimation of psycholinguistic norms using joint multidimensional annotations}

\name{Anil~Ramakrishna, Shrikanth~Narayanan}
\email{akramakr@usc.edu, shri@ee.usc.edu}
\address{Signal Analysis and Interpretation Laboratory, University of Southern California, \\ Los Angeles, CA}

\begin{document}

\maketitle

\begin{abstract}
    Psycholinguistic normatives represent various affective and mental constructs using numeric scores and are used in a variety of applications in natural language processing. They are commonly used at the sentence level, the scores of which are estimated by extrapolating word level scores using simple aggregation strategies, which may not always be optimal. In this work, we present a novel approach to estimate the psycholinguistic norms at sentence level. We apply a multidimensional annotation fusion model on annotations at the word level to estimate a parameter which captures relationships between different norms. We then use this parameter at sentence level to estimate the norms. We evaluate our approach by predicting sentence level scores for various normative dimensions and compare with standard word aggregation schemes.
\end{abstract}

\noindent\textbf{Index Terms}: Psycholinguistic normatives, Annotation fusion, Multidimensional annotations.


\section{Introduction}
Psycholinguistic norms are numeric ratings assigned to linguistic cues such as words or sentences to measure various psychological constructs. Examples include dimensions such as valence, arousal, and dominance which are used to analyze the affective state of the author (of the spoken or written text), along with norms of higher order mental constructs such as concreteness and imagability which have been associated with improvements in learning \cite{paivio1968concreteness}. The ease of computing the norms has enabled their application in a variety of tasks in natural language processing such as  information retrieval \cite{tanaka2013estimating}, sentiment analysis \cite{nielsen2011new}, text based personality prediction \cite{mairesse2007using} and opinion mining. The norms are typically annotated at the word level by psychologists who provide numeric scores to a curated list of seed words, which are then extrapolated to a larger vocabulary using either semantic relationships such as synonymy and hyponymy or using word occurrence based contextual similarity \cite{malandrakis2015therapy}. 

Most applications of psycholinguistic norms in NLP use sentence or document level scores, but manual annotation of the norms at these levels is difficult and not straightforward to generalize. In these cases, estimation of sentence level norms is done by aggregating the word level scores using simple averaging \cite{ramakrishna2017linguistic, malandrakis2015therapy}, or by using distribution statistics of the word level scores \cite{gibson2015predicting}. However, such strategies do not account for the non-trivial dependencies of sentence level semantics on the words, and may not be accurate at estimating the norms at the sentence level. In this work, we propose a new approach to estimate sentence level norms using inferred relationships between different dimensions along with partial annotations of the sentence level norms.

Annotation of the normatives at the sentence level is a challenging task when compared to word level annotations since it involves evaluating the underlying semantics of the sentence in the abstract space of the corresponding dimension, with some dimensions in particular being more difficult than others. For example, \textit{imagability}, a measure of how easy it is to create a mental image of the input word or sentence, is more difficult to annotate at the sentence level when compared to words. 
On the other hand, norms such as valence are relatively easier to annotate even at the sentence level in comparison. We use this observation along with the parameters learned from a joint annotation fusion model at word level to predict norms at sentence level.

Annotations are typically performed online using crowdsourcing platforms such as Amazon Mechanical Turk\footnote{mturk.com} (Mturk), which connect researchers with inexpensive workers from across the globe and provide easy scalability. 
Annotations are collected from several workers over a large number of instances, often on several related dimensions. These are then combined to obtain estimates for the label of interest, typically using aggregation techniques such as simple averaging or majority voting, or using more nuanced aggregation models which assume a structure for the annotators' behavior \cite{raykar2010learning}. The annotation dimensions are usually modeled independent of each other, but a few recent publications have explored joint modeling of the dimensions and have highlighted the benefits of this approach \cite{Ramakrishna2016, ramakrishna2020joint}. These models assume a joint relationship between the dimensions being annotated, and estimate model parameters that capture this relationship for each annotator, which can be used in estimating the sentence level normatives. Specifically, we can use model parameters learned at the word level to estimate the norms at the sentence level using partial sentence level annotations.

We use the model presented in \cite{ramakrishna2020joint}, in which the authors assume a matrix factorization model to capture the annotators' behavior, in which the annotations are assumed to be based on a linear transformation of the underlying label vector. Parameters of this model include a linear transformation matrix, $F_k$, which captures the individual contributions of each dimension in the annotation output. In our work, we assume that the annotator specific relationships between the dimensions captured by the parameter $F_k$ is comparable at both word and sentence levels. 
We collect word level annotations on valence, arousal and dominance and train the joint global annotation model from \cite{ramakrishna2020joint} to estimate the annotator parameters including $F_k$; we then use the word level estimates for $F_k$ on sentence level ratings from the same set of annotators. To predict sentence level scores of a given normative dimension, we make use of partial annotations on the remaining dimensions along with $F_k$. Our proposed approach shows improved performance in predicting the sentence level norms when compared to various word level aggregation strategies.

The rest of the paper is organized as follows. In Section \ref{sec: model}, we expand on the joint multidimensional annotation model and detail our data annotation approach in Section \ref{sec: data}, followed by experiments in Section \ref{sec: expts} and results in Section \ref{sec: results} before concluding in Section \ref{sec: conc}.

\begin{figure}
\centering
\includegraphics[trim={0.5cm 0cm 0cm 0cm},scale=0.45]{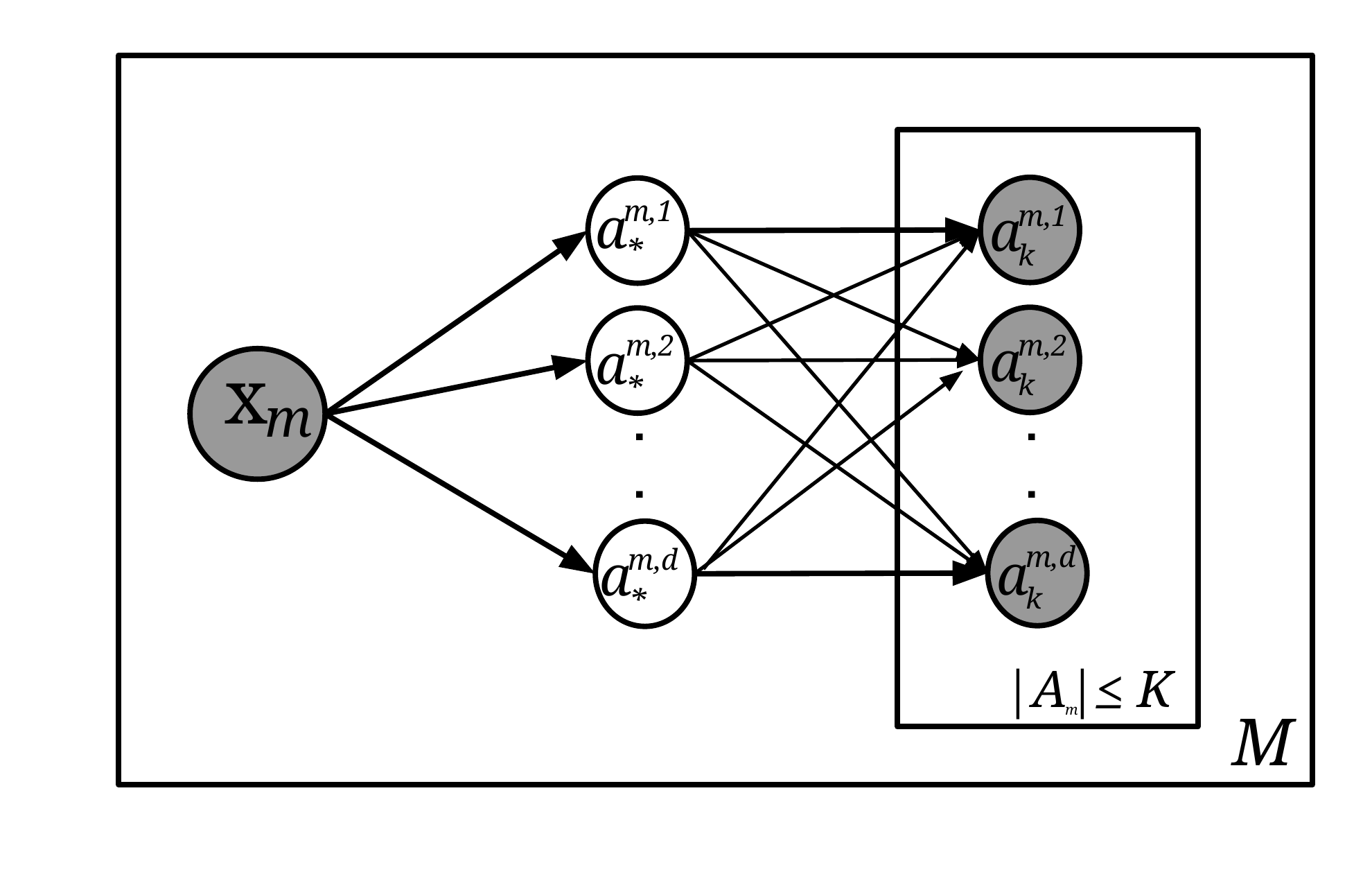}
\vspace{-20pt}
\caption{Proposed model. $\textbf{x}^m$ is the set of features for the $m^\text{th}$ data point, $a^{m,d}_*$ is the latent label for the $d^{th}$ dimension and $a^{m,d}_k$ is the rating provided by the $k^\text{th}$ annotator. Vectors $\textbf{x}^m$ and $\textbf{a}^m_k$ (shaded) are observed variables, while $\textbf{a}^m_*$ is latent. $\textbf{A}_m$ is the set of annotator ratings for the $m^\text{th}$ instance.}
\vspace{-5pt}
\label{fig:modelc}
\end{figure}

\section{Joint multidimensional annotation model}
\label{sec: model}
The annotation model is represented in plate notation in Figure \ref{fig:modelc}. In this model, the underlying label vector $\textbf{a}^m_*$ for each data instance is defined as a linear regression model as shown in Equation \ref{eqn:a_star_discrete}. An annotator, indexed by $k$, is assumed to apply a linear transformation function on vector $\textbf{a}^m_*$ to produce the annotation vector $\textbf{a}^m_k$ using the matrix $F_k$ as shown in Equation \ref{eqn:ann_discrete}.
\begin{align}
\label{eqn:a_star_discrete}
\textbf{a}^m_* &= \Theta^T\textbf{x}_m + \bm{\epsilon}_m \\
\label{eqn:ann_discrete}
\textbf{a}^m_k &= F_k\textbf{a}^m_* + \bm{\eta}_k 
\end{align}
where, $\textbf{x}_m \in \rm I\!R^P$; $\Theta \in  \rm I\!R^{P \times D}$; $\bm{\epsilon}_m \sim N(\textbf{0}, \sigma^2I)$; $\sigma^2 \in \rm I\!R$; $\bm{\eta}_k \sim N(\textbf{0}, \tau_k^2I)$;  $\tau_k^2 \in \rm I\!R$. $F_k \in  \rm I\!R^{D \times D}$  is the annotator specific linear transformation matrix. Each annotation dimension value $a^{m,d}_k$ for annotator $k$ is defined as a weighted average of the vector $\textbf{a}^m_*$ with weights given by $F_k(d,:)$.

In this model, the feature vector $\textbf{x}^m$ corresponding to each instance is assumed to be available, along with the annotations $\textbf{a}^m_k$, while the label vectors $\textbf{a}^m_*$ are assumed hidden, as shown in the Figure \ref{fig:modelc}. 
We use the EM algorithm from \cite{ramakrishna2020joint} to estimate the parameters, listed below for ease of exposition. Detailed derivations for the update equations below can be found in \cite{ramakrishna2020joint}.

We use Maximum Likelihood Estimation (MLE) to estimate the model parameters, in which we maximize the model likelihood shown below in Equation \ref{eqn:model_likelihood}.
\begin{align}
\log \mathcal{L} &= \sum_{m=1}^M \log p(\textbf{a}^m_1\dots\textbf{a}^m_K; \Phi) \nonumber \\
&= \sum_{m=1}^M \log \int_{\textbf{a}^m_*} p(\textbf{a}^m_1\dots\textbf{a}^m_K | \textbf{a}^m_*; F_k, \tau_k^2) p(\textbf{a}^m_*; \Theta,\sigma^2)  \, d\textbf{a}^m_* 
\label{eqn:model_likelihood}
\end{align}

Optimizing the above objective is non-trivial due to the presence of the integral within the log function. To address this, we use the well known Expectation Maximization algorithm \cite{dempster1977maximum}, which uses Jensen's inequality to derive a lower bound (shown below in Equation \ref{eqn:q_function}) on the objective based on current parameter estimates, by computing the expectation with respect to the conditional distribution $p(\textbf{a}^m_* | \textbf{a}^m_1\dots\textbf{a}^m_K)$.

\begin{align}
\log \mathcal{L} &= \sum_{m=1}^M \E_{\textbf{a}^m_* | \textbf{a}^m_1\dots\textbf{a}^m_K} \big[ \log \frac{p(\textbf{a}^m_1\dots\textbf{a}^m_K | \textbf{a}^m_*) p(\textbf{a}^m_*)}{q(\textbf{a}^m_*)}  \big]
\label{eqn:q_function}
\end{align}

This is followed by parameter estimation using maximization. The alternating expectation and maximization steps form the iterations of the EM algorithm. 

\begin{figure*}[ht]
    \centering
    \begin{subfigure}{0.48\textwidth}
        \centering
        \includegraphics[scale=0.5]{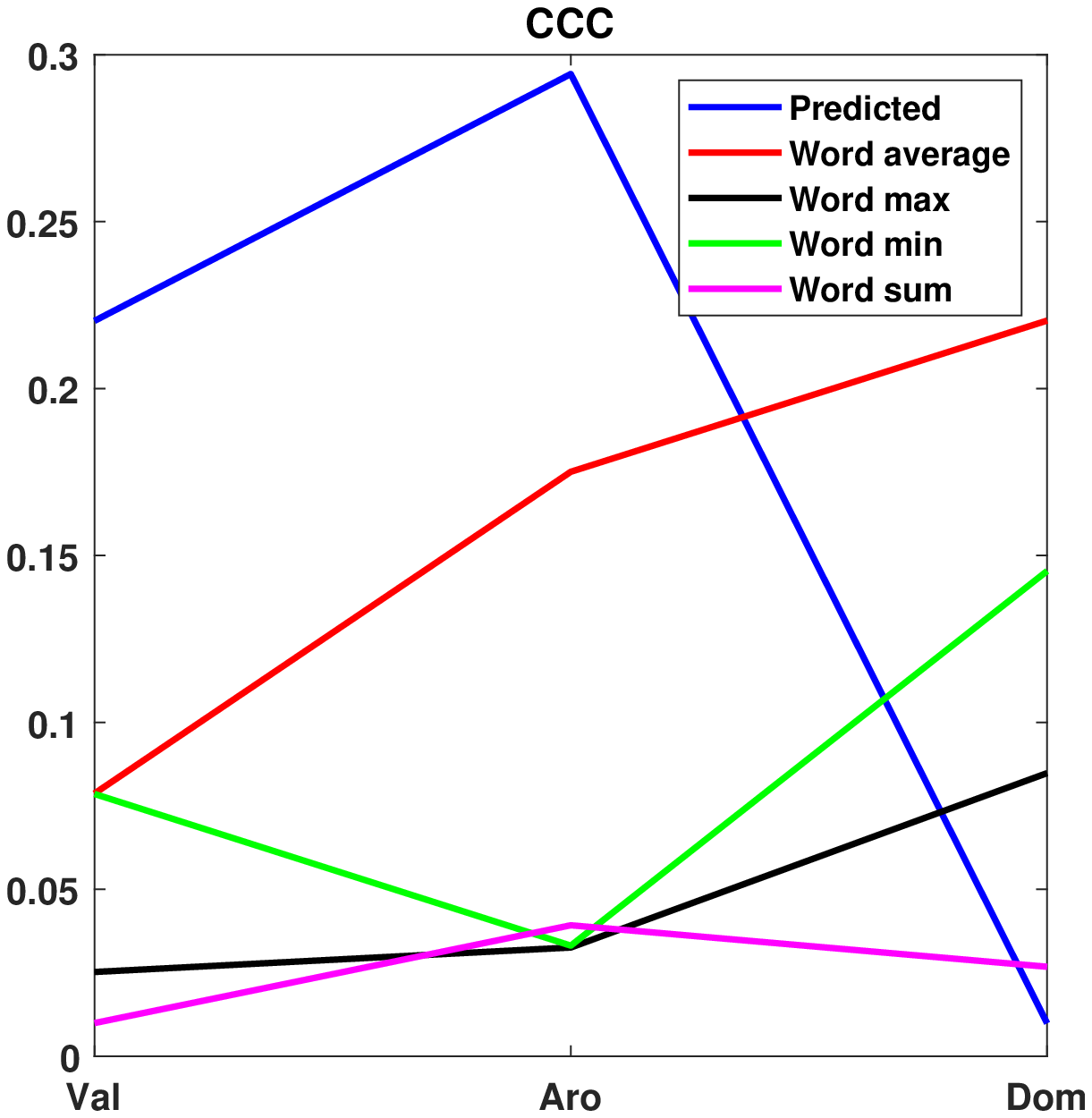}
        \caption{CCC}
        \label{fig:psyc_results_a}
    \end{subfigure}~
    \begin{subfigure}{0.48\textwidth}
        \centering
        \includegraphics[scale=0.5]{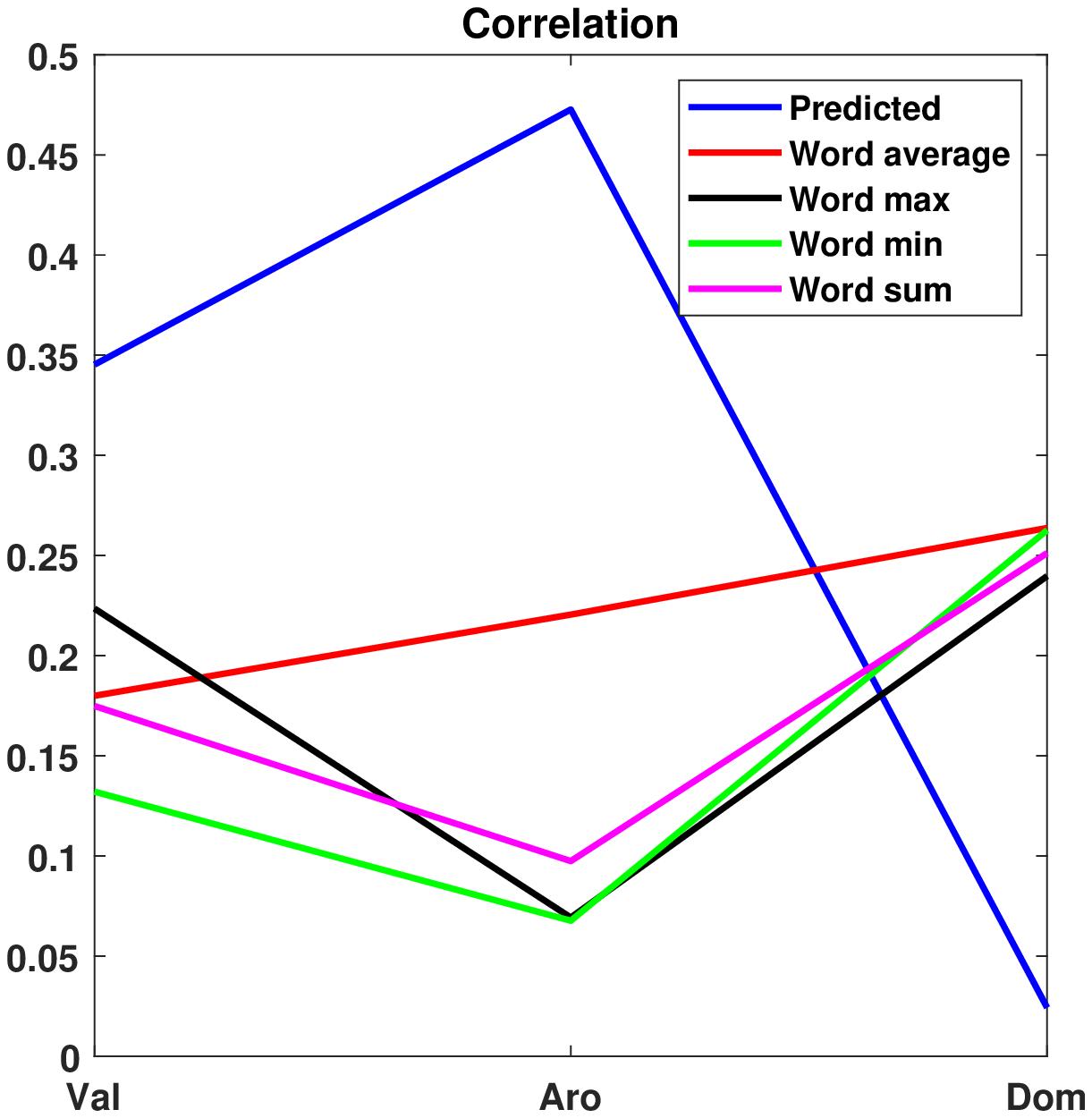}
        \caption{Pearson's Correlation}
        \label{fig:psyc_results_b}
    \end{subfigure}
    \caption{Performance of proposed and baseline models in predicting sentence level norms. Results show Concordance Correlation Coefficient (CCC) and Pearson Correlation values between the various estimates and the reference expert ratings on the EmoBank corpus. The estimates of the proposed model for Valence and Arousal are superior while those for Dominance are poor; subsequent analysis show poor human interannotator agreement for dominance ratings as a possible reason. See also Figure 3.}
    \label{fig:psyc_results}
\end{figure*}

\subsection{EM algorithm}
\label{sec: em}
\textbf{Initialization} The model is initialized by assigning the mean of annotations for each data instance as the estimate for $\textbf{a}^m_*$. Given this, the initial parameters are estimated using update equations listed in the maximization step below. 

\textbf{E-step} We compute the expected value of $\textbf{a}^m_*$ with respect to the distribution $p(\textbf{a}^m_* | \textbf{a}^m_1\dots\textbf{a}^m_K)$, which is assumed to be it's \textit{soft} estimate for each data instance. 
\begin{gather}
\E_{\textbf{a}^m_* | \textbf{a}^m_1\dots\textbf{a}^m_K}[\textbf{a}^m_*] = \Theta^T \textbf{x}_m +  \Sigma_{\textbf{a}^m_*,\textbf{a}^m_1\dots\textbf{a}^m_K} \Sigma^{-1}_{\textbf{a}^m_1\dots\textbf{a}^m_K,\textbf{a}^m_1\dots\textbf{a}^m_K} \nonumber \\(\textbf{a}^m - \boldsymbol{\mu}^m) \nonumber \\
\Sigma_{\textbf{a}^m_* | \textbf{a}^m_1\dots\textbf{a}^m_K} [\textbf{a}^m_*] = \Sigma_{\textbf{a}^m_*,\textbf{a}^m_*} - \Sigma_{\textbf{a}^m_*,\textbf{a}^m_1\dots\textbf{a}^m_K}\Sigma^{-1}_{\textbf{a}^m_1\dots\textbf{a}^m_K,\textbf{a}^m_1\dots\textbf{a}^m_K} \nonumber \\ \Sigma_{\textbf{a}^m_1\dots\textbf{a}^m_K,\textbf{a}^m_*} \nonumber
\end{gather}

\textbf{M-step} Given the soft estimate for $\textbf{a}^m_*$, parameter estimates are computed by maximizing Equation \ref{eqn:q_function}. The update equations for this step are listed below.
\begin{gather}
\Theta = (\textrm{X}^T\textrm{X})^{-1}(\textrm{X}^T\E[\textrm{a}^m_*]) \nonumber
\end{gather}
\vspace{-0.8cm}
\begin{gather}
F_k = \bigg( \sum_{m=1}^{M_k} \textbf{a}^m_K \E[\left(\textbf{a}^m_*\right)^T] \bigg) \bigg( \sum_{m=1}^{M_k}  \E[\textbf{a}^m_*\left(\textbf{a}^m_*\right)^T]\bigg)^{-1} \nonumber
\end{gather}
\vspace{-0.5cm}
\begin{gather}
\sigma^2 = \frac{1}{md} \sum_{m=1}^{M} \bigg( \E[(\textbf{a}^m_*)^T\textbf{a}^m_*] -   2tr\left(\Theta'^T\textbf{x}_m\E[(\textbf{a}^m_*)^T]\right) \nonumber \\ + tr(\textbf{x}^T_m\Theta'\Theta'^T\textbf{x}_m) \bigg) \nonumber 
\end{gather}
\vspace{-0.8cm}
\begin{gather}
\tau_k^2 = \frac{1}{{m_k}d} \sum_{m=1}^{M_k} \bigg( (\textbf{a}^m_K)^T\textbf{a}^m_K - 2tr\big(F_k'^T \textbf{a}^m_K\E[(\textbf{a}^m_*)^T]\big) \nonumber \\ + tr\big(F_k'^TF_k'\E[\textbf{a}^m_*(\textbf{a}^m_*)^T]\big)  \bigg) \nonumber
\end{gather}

\textbf{Termination} We terminate the algorithm when the change in model log-likelihood reduces to less than $0.001 \%$ from the previous iteration. 

\section{Data annotation}
\label{sec: data}

We performed two sets of experiments, collecting word and sentence level annotations on specific dimensions in each. In the first experiment (which we refer to as \textit{VAD} from now), we collected annotations on the affective norms of Valence, Arousal and Dominance using Mturk for words sampled from \cite{warriner2013norms}. This corpus was chosen because it provides expert ratings on Valence, Arousal and Dominance for nearly 14,000 English words. Annotators were asked to provide numeric ratings between 1 to 5 (inclusive) for each dimension, on assignments consisting of a set of 20 words. In total, we collected 20 annotations each on a set of 200 words randomly sampled from \cite{warriner2013norms}. Instructions for the annotation assignments included definitions along with examples for each of the dimensions being annotated. After filtering incomplete and noisy submissions, we retained only those annotators who provided ratings for at least 100 words in the subsequent sentence level annotation task, to ensure sufficient training data.

Sentence level annotations were collected on sentences from the Emobank corpus \cite{buechel2017emobank}, which includes expert ratings on valence, arousal and dominance for 10000 English sentences. 21 different annotators from the word level annotation task described above were invited to provide labels for 100 sentences randomly sampled from this corpus. The assignments were presented in a similar fashion as word level annotations, with each assignment including 10 sentences and the workers providing numeric ratings for valence, arousal and dominance for each sentence. We use the annotator specific parameters $F_k$ estimated at the word level to predict the norms at sentence level using the approach described in the next section. 

In our second experiment (which we refer to as \textit{IGP} from now), we collected word and sentence level annotations on three new psycholinguistic normative dimensions: imagability, which measures the degree of the stimulus' proclivity to create a mental picture;  genderladenness, which measures the degree of masculine or feminine association evoked by the stimulus; and pleasantness, which measures the degree of pleasant feelings associated with the stimulus. We used the same words and sentences used in our previous experiment for annotations on valence, arousal and dominance. Since we do not have expert ratings for pleasantness, imagability and genderladenness, we use the strategy followed in AVEC 2018 challenge to evaluate model performance. 

\section{Experiments}
\label{sec: expts}
Given annotator parameters $F_k$ estimated at the word level, we use partial annotator ratings at the sentence level to predict the remaining norms. For example, in the \textit{VAD} experiment, while predicting sentence level scores of valence, we use the sentence level annotator ratings on arousal and dominance along with the word level parameter matrix $F^{\text{word}}_k$. The use of partial annotations enables us to predict sentence level norms on challenging psycholinguistic dimensions using ratings on dimensions which maybe easier to annotate. 

\begin{figure}
    \centering
    \begin{subfigure}{0.4\linewidth}
        \centering
        \includegraphics[scale=0.35]{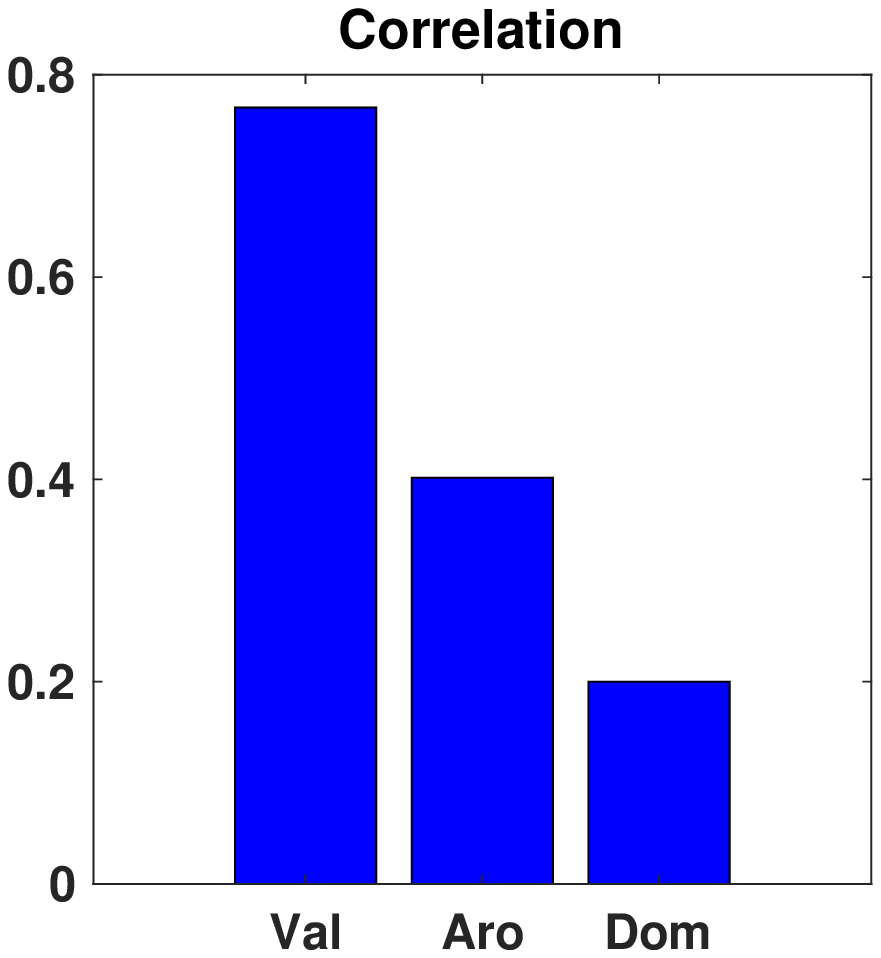}
        \caption{Pearson's Correlation}
    \end{subfigure}~\hspace{0.5cm}
    \begin{subfigure}{0.4\linewidth}
        \centering
        \includegraphics[scale=0.35]{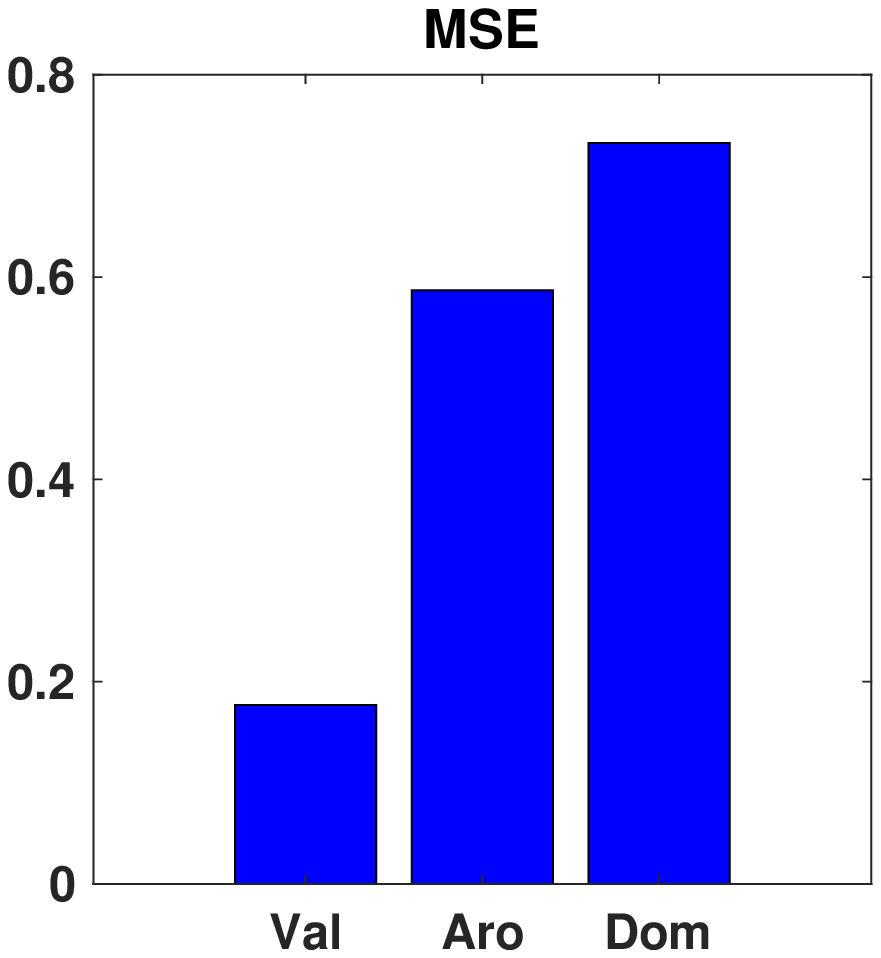}
        \caption{Mean Squared Error}
    \end{subfigure}
    \caption{Performance of the best annotators for each dimension (but over all instances) in our dataset and annotator average when compared with expert ratings from the Emobank corpus}
    \label{fig:psyc_best_results}
\end{figure}

\begin{figure*}
    \centering
    \begin{subfigure}[b]{0.3\textwidth}
        \centering
        \includegraphics[scale=0.45]{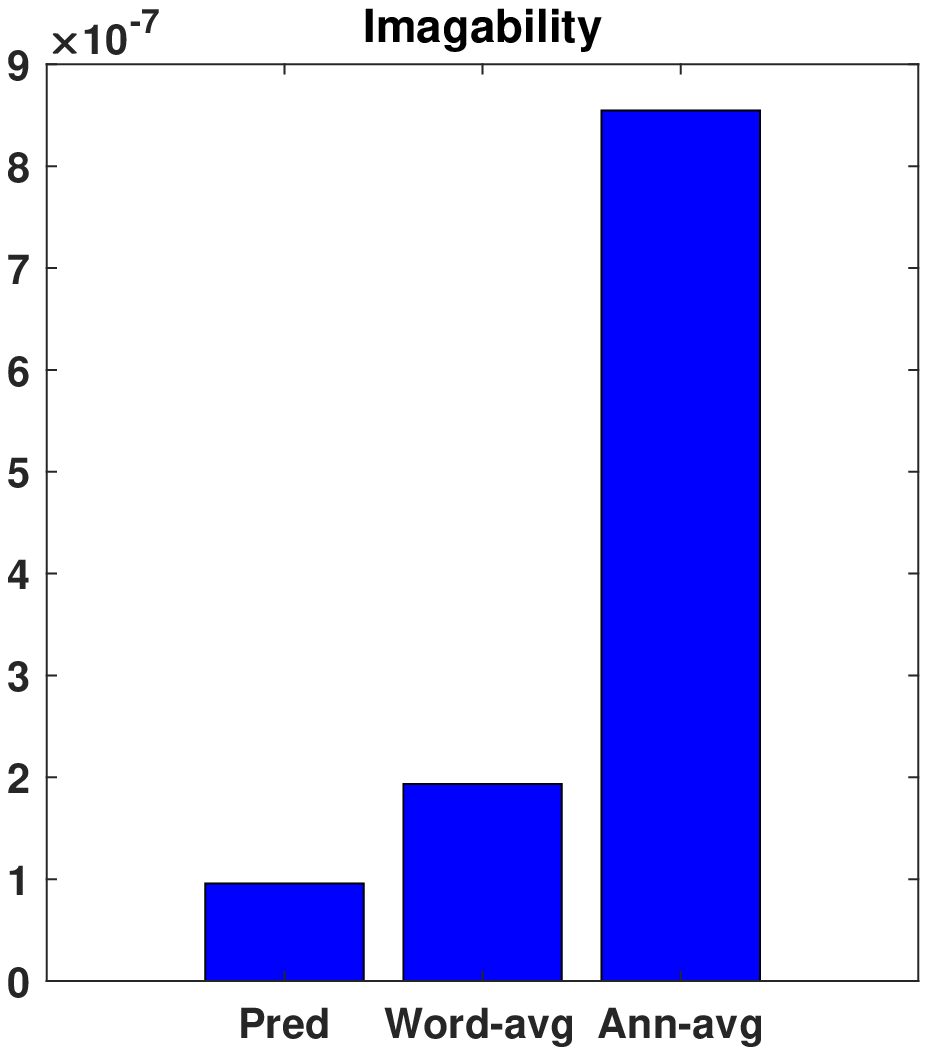}
        \caption{Imagability}
    \end{subfigure}~
    \begin{subfigure}[b]{0.3\textwidth}
        \centering
        \includegraphics[scale=0.45]{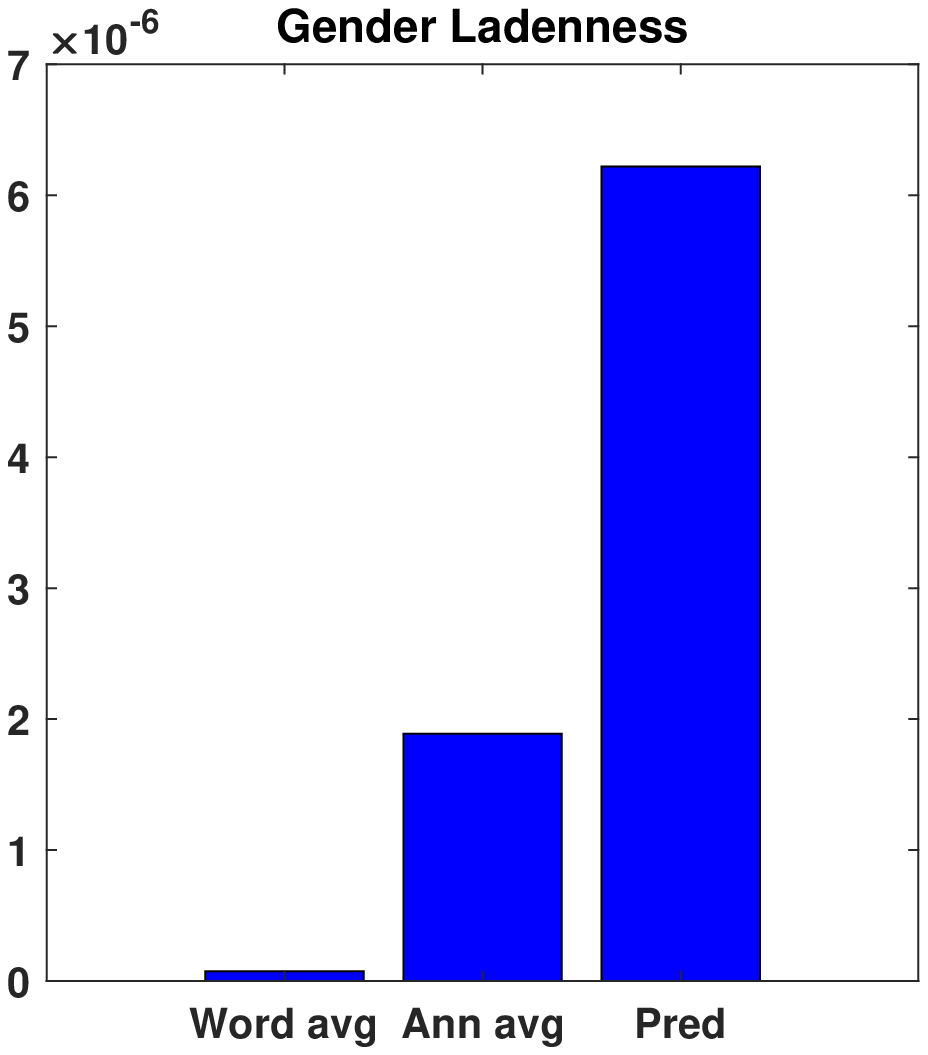}
        \caption{Genderladenness}
    \end{subfigure}~
    \begin{subfigure}[b]{0.3\textwidth}
        \centering
        \includegraphics[scale=0.45]{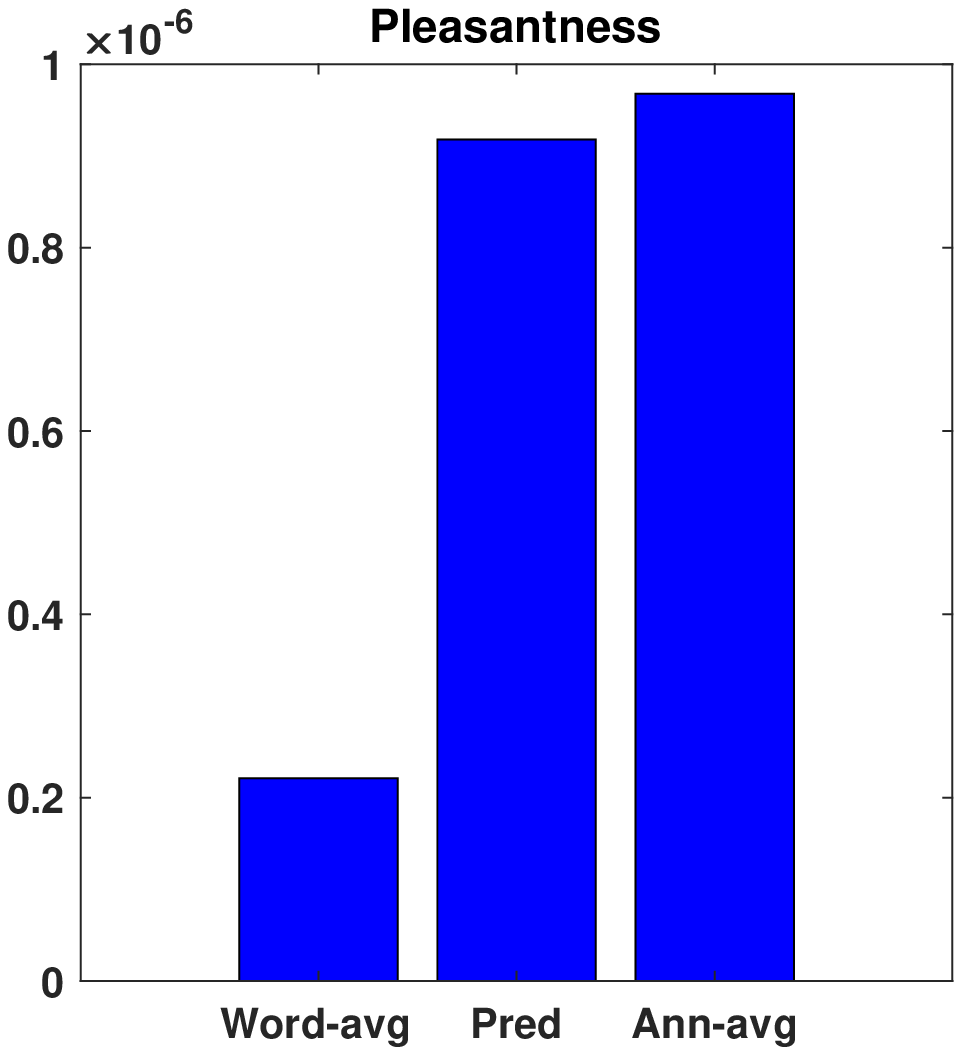}
        \caption{Pleasantness}
    \end{subfigure}
    \caption{MSE of predicted and baseline models in predicting Imagability, Genderladenness and Pleasantness}
    \label{fig:gip_plots}
\end{figure*}

\begin{align}
        \label{eqn:normpred}
        \centering
        &\begin{bmatrix}
            \cdot \\
            a^{m, d'}_1 \\
            \vdots \\
            a^{m, d'}_K \\
            \cdot
        \end{bmatrix}
        =
        \begin{bmatrix}
            \cdot \\
            F^{\text{word}}_1[d',:] \\
            \vdots \\
            F^{\text{word}}_K[d',:] \\
            \cdot
        \end{bmatrix}
        \begin{bmatrix}
            \\ 
            \\
            \textbf{a}^m_* \\
            \\
            \\
        \end{bmatrix}
\end{align}

\noindent where, $d' \in \{1 \dots D\}; d' \neq d$  is the dimension to predict. $\textbf{a}^m_*$ is estimated using linear regression.

In both our experiments, we make use of the IID Gaussian noise assumption in Equation \ref{eqn:ann_discrete}, which reduces the task of predicting the sentence level norm to a linear regression problem shown in Equation \ref{eqn:normpred}. Rows of the matrix $F^{\text{word}}_k$ are treated as features of the regression model with vector $\textbf{a}^m_*$ as the regression parameter. Given sentence level partial annotations $\textbf{a}^{m,\setminus d}_k$ (vector $\textbf{a}^{m_k}$ with $a_k^{m,d}$ removed), and matrix $F^{\text{word}}_k$, the regression parameter vector $\textbf{a}^m_*$ can be estimated using normal equations or gradient descent. For each dimension within a given experiment, we use Equation \ref{eqn:normpred} to estimate the sentence level normatives. 
The features $x_m$ used in both our experiments were 300 dimensional GloVe embeddings \cite{pennington2014glove} at word level, which were aggregated using simple averaging at sentence level.  

In the \textit{VAD} experiment, we compare the predicted dimensions with expert ratings from the Emobank corpus, which acts as our reference to evaluate model performance. For baselines, we compute different aggregations of word level normative scores after filtering out non-content words as is common in literature \cite{ramakrishna2017linguistic}. Word level scores for the norms were computed using the approach described in \cite{malandrakis2015therapy}. We used unweighted average, maximum, minimum and sum of the word level norms as the baseline aggregation functions. 

In the \textit{IGP} experiment, we train linear regression models using predictions from the proposed model and directly compare the training set error with baselines. Low training error implies \textit{higher learnability} (due to better correlations with the features) of the predicted signal and serves as a crude proxy for quality.
For baselines, we use training error from labels obtained by simple averaging of word level normative scores, and sentence level average of annotations. 

We use Concordance Correlation Coefficient ($\rho_c$) \cite{lawrence1989concordance} and the Pearson's correlation coefficient ($\rho$) as evaluation metrics. $\rho_c$ measures any departures from the \textit{concordance line} (line passing through the origin at $45\degree$ angle). Hence it is sensitive to rotations or rescaling in the predicted values of $\textbf{a}^m_*$. Given two samples $x$ and $y$, the sample concordance coefficient $\hat{\rho_c}$ is defined as shown below.
\begin{equation}
\label{eqn:concordance_correlation}
\hat{\rho_c} = \frac{2s_{xy}}{s^2_x+s^2_y+(\bar{x} - \bar{y})^2}
\end{equation}

\noindent where $s_x$ and $s_y$ are sample standard deviations, while $s_{xy}$ is the sample covariance.

\vspace{-0.3cm}
\section{Results}
\label{sec: results}

\vspace{-0.2cm}
\subsection{\textit{VAD}}
Figure \ref{fig:psyc_results} shows the performance of the proposed model along with the different baselines. As seen from the figure, the proposed model outperforms the baselines in predicting valence and arousal in both evaluation metrics, suggesting the efficacy of the approach. 
Using partial ratings at sentence level along with matrix $F_k$ which captures relationships between the dimensions, the proposed approach seems to outperform the baseline word aggregation schemes in these two dimensions. 
Performance in $\rho_c$ appears to be lower than $\rho$, suggesting the presence of a rotation in the predicted values. This can be attributed to the unidentifiability commonly observed in matrix factorization models such as the annotation fusion model of \cite{ramakrishna2020joint}. Common solutions to address this involve assuming a suitable prior on the parameter $F_k$, which may lead to better estimates of $\rho_c$.

Model performance on dominance, on the other hand, is considerably low in both metrics. To further investigate the reason for this, we examined the performance of the best possible annotator for each dimension in this experiment and compare their predictions with the expert ratings from the Emobank corpus in Figure \ref{fig:psyc_best_results}.
Evidently, for dominance, we notice very low correlation and high MSE between our best annotators and the experts, suggesting a high disagreement for this dimension. This may have been due to a possibly differing definition and/or interpretation of dominance between the two sets of annotators. 

\vspace{-0.2cm}
\subsection{\textit{IGP}}
In our second experiment, we use model training error as a proxy for evaluating prediction quality since we do not have expert ratings. Figure \ref{fig:gip_plots} shows the training error for the proposed model when compared with two baselines. The proposed model shows lowest training error in predicting imagability while the performance is relatively worse in genderladenness and pleasantness, suggesting relatively stronger dependency of imagability on the other dimensions.

\section{Conclusion}
\label{sec: conc}
We presented a novel computational approach to estimate sentence level psycholinguistic norms using joint multidimensional annotation fusion. 
We evaluate our approach by predicting sentence level normatives on various dimensions in two different experiments, and showed improvements in specific cases. 
Future work includes evaluating the model on more abstract psycholinguistic dimensions such as concreteness and dominance. 
The primary challenge there lies in obtaining expert ratings on these dimensions at the sentence level to evaluate the model predictions. Recently, alternate schemes to evaluate a model in the absence of a reliable ground truth or reference have been proposed, such as the evaluation strategy used in the AVEC 2018 challenge \cite{ringeval2018avec}. The challenge organizers proposed a scheme where annotation fusion models are evaluated by training regression models on labels predicted by the fusion models which are evaluated on a disjoint test set. 

\bibliographystyle{IEEEtran}
\bibliography{references.bib}

\end{document}